\pdfoutput=1

\documentclass[11pt]{article}
\usepackage{graphicx}
\usepackage{tipa}
\usepackage{colortbl}
\usepackage{xcolor,color}

\usepackage{acl}

\usepackage{times}
\usepackage{latexsym}

\usepackage[T1]{fontenc}

\usepackage[utf8]{inputenc}

\usepackage{microtype}

\title{A New Framework for Fast Automated Phonological Reconstruction Using Trimmed Alignments and Sound Correspondence Patterns}

\author{Johann-Mattis List \\
  DLCE \\
  MPI-EVA \\
  Leipzig\\
  {\footnotesize\texttt{mattis\_list@eva.mpg.de}} \\\And
  Robert Forkel \\
  DLCE \\
  MPI-EVA \\
  Leipzig\\
  {\footnotesize\texttt{robert\_forkel@eva.mpg.de}} \\\And
  Nathan W. Hill \\
  Trinity Centre for Asian Studies  \\
  University of Dublin \\
  Dublin\\
  {\footnotesize\texttt{nathan.hill@tcd.ie}}}

\begin{document}
\maketitle
\begin{abstract}
Computational approaches in historical linguistics have been increasingly applied during the past decade and many new methods that implement parts of the traditional comparative method have been proposed. Despite these increased efforts, there are not many easy-to-use and fast approaches for the task of phonological reconstruction. Here we present a new framework that combines state-of-the-art techniques for automated sequence comparison with novel techniques for phonetic alignment analysis and sound correspondence pattern detection to allow for the supervised reconstruction of word forms in ancestral languages. We test the method on a new dataset covering six groups from three different language families. The results show that our method yields promising results while at the same time being not only fast but also easy to apply and expand.
\end{abstract}

\section{Introduction}
Phonological reconstruction is a technique by which words in ancestral languages, which may not even be reflected in any sources, are restored through the systematic comparison of descendant words (\textit{cognates}) in descendant languages \citep{Fox1995}. Traditionally, scholars apply the technique manually, but along with the recent quantitative turn in historical linguistics, scholars have increasingly tried to automate the procedure. Recent automatic approaches for linguistic reconstruction, be they supervised or unsupervised, show two major problems. First, the underlying code is rarely made publicly available, which means that they cannot be further tested by applying them to new datasets. Second, the methods have so far only been tested on a small amount of data from a limited number of language families.  
Thus, \citet{Bouchard-Cote2013} report remarkable results on the reconstruction of Oceanic languages, but the source code has never been published, and the method was never tested on additional datasets.
\citet{Meloni2021} report very promising results for the automated reconstruction of Latin from
Romance languages, using a new test set derived from a dataset originally provided by
\citet{Dinu2014}, but they could only share part of the data, due to restrictions underlying the
data by \citet{Dinu2014}. \citet{Bodt2021} experiment with the prediction of so far unelicited words
in a small group of Sino-Tibetan languages, which they registered prior to verification
\citep{Bodt2019}, but they do not test the suitability of their approach for the reconstruction of ancestral languages. \citet{Jaeger2019} presents a complete pipeline by which words are clustered into cognate sets and ancestral word forms are reconstructed, but the method is only tested on a very small dataset of Romance languages.
 
With increasing efforts to unify and standardize lexical datasets from different sources \citep{Forkel2018a}, more and more datasets that could be used to test methods for automated linguistic reconstruction have become available. Additionally, thanks to the huge progress which techniques for automated sequence comparison have made in the past decades \citep{Kondrak2000,Steiner2011,List2014d}, it is much easier today to combine existing methods into new frameworks that tackle individual tasks in computational historical linguistics.
 
In this study, we present a new framework for automated linguistic reconstruction which combines state-of-the-art methods for automated sequence comparison with fast machine-learning techniques and test it on a newly compiled test set that covers multiple language families. 

\begin{table}[h!]
    \centering
    \resizebox{\linewidth}{!}{%
    \begin{tabular}{|l|l|l|c|c|c|}\hline
        Name & Source & Subgroup & L & C & W \\\hline\hline

        \texttt{Bai}       & Wang (2004)          & Bai       & 10 & 459 & 3866 \\
        \texttt{*Burmish}  & Gong and Hill (2020) & Burmish   & 9  & 269 & 1711 \\
        \texttt{*Karen}    & Luangthongkum (2020) & Karen     & 11 & 365 & 3231 \\
        \texttt{Lalo}      & Yang (2011)          & Lalo (Yi) & 8  & 1251 & 7815 \\
        \texttt{Purus}     & Carvalho (2020)      & Purus     & 4  & 199 & 693 \\
        \texttt{Romance}   & Meloni et al. (2021) & Romance   & 6  & 4147 & 18806 \\
	\hline
    \end{tabular}}
    \caption{Datasets used in this study (L=Languages, C=Cognate Sets, W=Word Forms *=new data prepared for this study).}
    \label{tb1}
\end{table}
\begin{figure*}[t!]
\centering
\includegraphics[width=\textwidth]{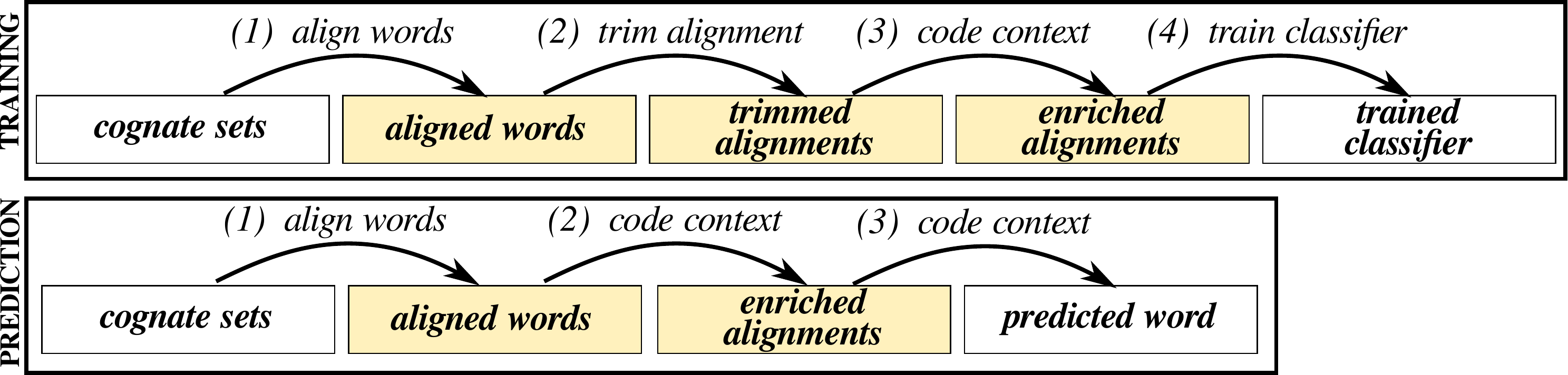}
\caption{Workflow for the new framework for word prediction and linguistic reconstruction based on gap-free alignments and sound correspondence patterns.}\label{fig1}
\end{figure*}
\section{Materials}
The number of cross-linguistic datasets amenable for automated processing has been constantly increasing during the past years, as reflected specifically also in the development of standards for data representation that are increasingly used by scholars (see \citealt{Forkel2018a} as well as \citealt{List2021PREPRINTd} for recent initiatives to make standardized cross-linguistic wordlists available in the form of open repositories). Unfortunately, the number of datasets in which proto-languages are provided along with descendant languages is still rather small. 
For the experiments reported here, a new cross-linguistic collection of six datasets from three
language families (Sino-Tibetan, Purus, and Indo-European) was created. Datasets were all taken from
published studies and then converted to Cross-Linguistic Data Formats (CLDF) \citep{Forkel2018a} using the
CLDFBench Python package \citep{Forkel2020} with the \texttt{PyLexibank} plugin
\citep{PyLexibank}. 
 
CLDF allows for a consistent handling of data when using
software like Python or R. In addition, CLDF offers several levels of standardization by allowing to
link the data to existing reference catalogs, such as Glottolog \citep{Glottolog} for languages,
Concepticon for concepts \citep{Concepticon}, or Cross-Linguistic Transcription Systems
\citep{Anderson2018,CLTS} for speech sounds. 

While three of the datasets (Bai, Lalo, and Purus) had been previously included into the Lexibank
collection, a repository of lexical datasets in Cross-Linguistic Data Formats
\citep{List2021PREPRINTd}, we converted the open part of the Latin dataset by
\citet{Meloni2021} to CLDF. Additionally, we converted a selection of a smaller part of the data by
\citet{Gong2020} to CLDF and retro-standardized the data by \citet{Luangthongkum2019}.
While all datasets provided forms for ancestral languages, not all datasets provided the direct links
between these proto-forms and the reflexes in the descendant languages in the form of annotations
indicating cognacy. While these were added manually for the Karen data, using the EDICTOR tool for
etymological data curation \citep{List2017d,EDICTOR}, we used the automated method
for partial cognate detection by \citet{List2016g} to cluster proto-forms and reflexes into cognate
sets for the data on Bai, Lalo, and Purus. 
 
The datasets, along with their sources and some basic information regarding the number of languages (L), cognate sets (C), and word forms (W) are listed in Table \ref{tb1}.
The collection offers a rather diverse selection, in which the amount of data varies both with respect to the number of word forms, cognate sets, and languages.

\nocite{Gong2020,Yang2011,Wang2004b,Meloni2021,Luangthongkum2019,Carvalho2021}
\section{Methods}

\subsection{Workflow}
The new framework can be divided into a training and a prediction stage. The training consists of four steps. In step~(1), the cognate sets in the training data are \emph{aligned} with a multiple phonetic alignment algorithm. In step~(2), the alignments are \emph{trimmed} by merging sounds in the ancestral language into clusters which would leave no trace in the descendant languages (§~\ref{trim}). In step~(3), the alignments of the descendant languages are enriched by \emph{coding for context} that might condition sound changes (§~\ref{coding}). In step~(4) the enriched alignment sites are assembled and fed to a \emph{classifier} for training. 
 
The prediction consists of three steps. Given a cognate set as input, the word forms are aligned with the help of the same algorithm for multiple alignment used in the training phase in step~(1). In step~(2), the alignment is enriched using the same method applied in the training phase and then passed to the classifier to predict the word form in the ancestral language in step~(3).

Figure \ref{fig1} illustrates the workflow, which is flexible with respect to individual methods used for individual steps. For phonetic alignment, we use the Sound-Class-Based Phonetic Alignment (SCA) algorithm \citep{List2012c}, which is the current state-of-the-art method, but any other method that yields multiple alignments could be used. The same holds for the trimming procedure, (see §~\ref{trim}), the enrichment procedure, (see §~\ref{coding}), or the classifier (see §~\ref{classifiers}). 
\subsection{Trimming Alignments}\label{trim}

Using multiple alignments to predict ancestral or new words is nothing new and has essentially been practised by classical historical linguists for a long time \citep{Grimm1822}. That multiple alignments can also be used in computational frameworks has been demonstrated by \citet{List2019a}, who inferred correspondence patterns from phonetic alignments and later used these correspondence patterns to predict words missing from the data. One problem not considered in this approach, however, is that correspondence patterns can only be inferred for those cases in which descendant languages have a reflex for a given sound in the ancestral language. In those cases where the sound has been lost, a prediction is not possible. 

\begin{figure}[t!]
    \centering
    \includegraphics[height=3.75cm]{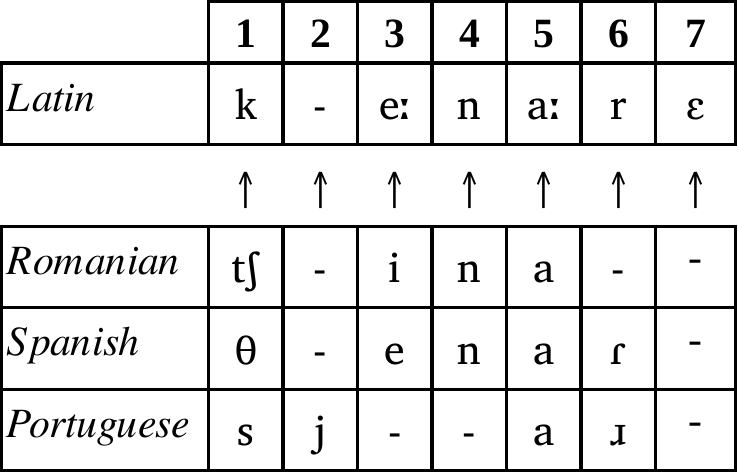}
    \caption{Prediction problems when ancestral segments in multiple alignments do not show reflexes in the descendant languages.}
    \label{fig:fig2}
\end{figure} 

\begin{figure}[b!]
    \centering
    \includegraphics[height=3.75cm]{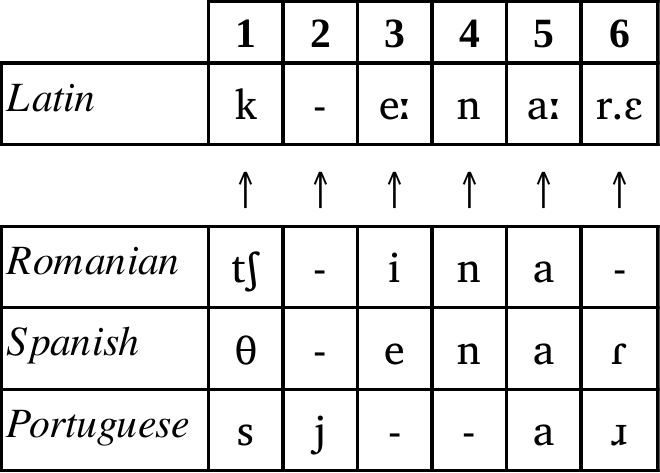}
    \caption{Trimming alignments by merging sounds in the ancestral languages in those cases where an alignment column does not have sound reflexes in the descendant languages.}
    \label{fig:fig3}
\end{figure}

This problem is illustrated in Figure \ref{fig:fig2}, where the Latin ending [\textipa{E}] has no
reflex sound in either of the descendant languages in the sample, yielding an alignment column that
is completely filled with gap symbols. Our solution to deal with this problem is to post-process the
multiple alignments in the training procedure by merging those columns which show only gaps in the
descendant languages with the preceding alignment column. This is illustrated in Figure
\ref{fig:fig3}, where the Latin ending is now represented as a single sound unit [\textipa{r.E}].
This trimming procedure, which was introduced for by \citet{Ciobanu2018} for pairwise alignments
and is here extended to multiple alignments, is justified by the fact that correspondence patterns
preceding lost sounds usually convey enough information to be distinguished from those patterns in
which no sound has been lost.   

\subsection{Coding Context}\label{coding}
\begin{figure}[t!]
    \centering
    \includegraphics[width=0.8\linewidth]{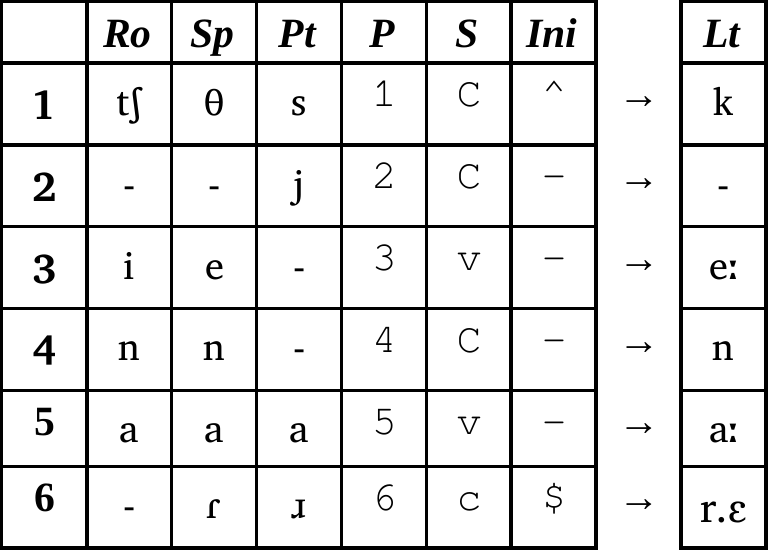}
    \caption{Enriching a phonetic alignment by coding various forms of context.}
    \label{fig:fig4}
\end{figure}

Previous alignment-based approaches to automated word prediction have made exclusive use of the
information provided by individual correspondence patterns derived from phonetic alignments
\citep{List2019a}. While this has shown to yield already surprisingly good results, we know well
that sound change often happens in certain phonetic environments. For example, we know that the
initial position of a word is typically much stronger and less prone to change than the final
position \citep{Geisler1992}. Similarly, consonants in the syllable onset position (preceding a
vowel) also tend to show different types of sound change compared to consonants in the syllable
offset \citep{List2014d}. Last but not least, certain sound changes may be due to ``long-range
dependencies'', or supra-segmental features like
tone, which is typically marked in the end of a morpheme in the phonetic transcription of South-East
Asian languages. In order to allow a classifier to make use of this information, our framework
allows to enrich the phonetic alignments further, by deriving contextual information from individual
phonetic alignments and adding it to the correspondence patterns that are then used to train the
classifier. An example for this procedure is given in Figure \ref{fig:fig4}, where the phonetic
alignment is given in transposed form (switching columns and rows), with each row corresponding to one correspondence pattern.
While the information from correspondence patterns alone would only account for the first three
columns of the matrix, three additional types of phonetic context have been added.  Thus, column
\textit{P}
indicates the position of a pattern in the form of an index.  
Column \textit{S} provides information on the syllable structure following \citet{List2014d}, and
column \textit{Ini} indicates,
whether a pattern occurs in the beginning (\verb|^|), the end (\verb|$|) or the middle
(\verb|-|) of a word form. Enriching alignments should be done in a careful way, in order to avoid
over-fitting the classifier. In our experiments, we contrast all eight possible combinations,
ranging from the full coding shown in Figure \ref{fig:fig4}, up to a coding of the alignment without additional enrichment.  

\subsection{Classifiers}\label{classifiers}
Our approach is very flexible with respect to the choice of the classifier. In order to keep the
approach \emph{fast}, we decided to restrict our experiments to the use of a Support Vector Machine
(SVM) with a linear kernel, since SVMs have been successfully applied in recent approaches in
computational historical linguistics dealing with different classification tasks
\citep{Jaeger2017,Cristea2021}. We compare this approach with the graph-based method based on
correspondence patterns (henceford called CorPaR) presented by \citet{List2019a}, which we modified
slightly. While the original method uses a greedy algorithm to identify the largest cliques in the
network, we now compute all cliques and rank them by counting the number of nodes they cover. 
An alignment site in an alignment is now compared against the consensus patterns extracted from the
cliques in the graph and the prediction for the pattern with the largest number of reflexes is taken
as the prediction. When no compatible pattern can be found, a search for the best candidates among
patterns that are only partially compatible with the alignment site is invoked. This increases the
chances too find a suitable reconstruction in those cases where the correspondence patterns are not fully
regular.

\subsection{Evaluation}
Most scholars tend to report only the edit distance -- also called Levenshtein distance
\citep{Levenshtein1965} -- between the predicted and the attested string, both normalized by the
length of the longer string and in unnormalized form. However, reporting the edit distance alone has
the disadvantage that systematic differences between predicted and attested forms may be penalized
too high, which is why we follow \citet{List2019e} in computing the \textit{B-Cubed F-scores}
\citep{Amigo2009} of the alignments of source and target sequences. B-Cubed F-Scores measure the difference
between two classifications, ranging from 0 to 1, with 1 indicating complete similarity with respect
to the structure of the classifications. Since the prediction of words can be seen as a
classification task in which a certain number of sound slots should be classified by rendering them
as identical or different from each other, B-Cubed F-Scores do not measure whether automated
reconstructions are identical with attested reconstructions in the gold standard, but rather whether
automated reconstructions approximate the structure of the reconstructions in the gold standard. As
a result, B-Cubed F-Scores can show to which degree an automated reconstruction comes structurally
close to the gold standard, even if individual reconstructed sounds differ. Given that B-Cubed
F-Scores measure consistency across a set of reconstructed word forms, they should not be applied to
individual items.

\begin{figure*}[t!]
    \centering
    \includegraphics[width=0.75\textwidth]{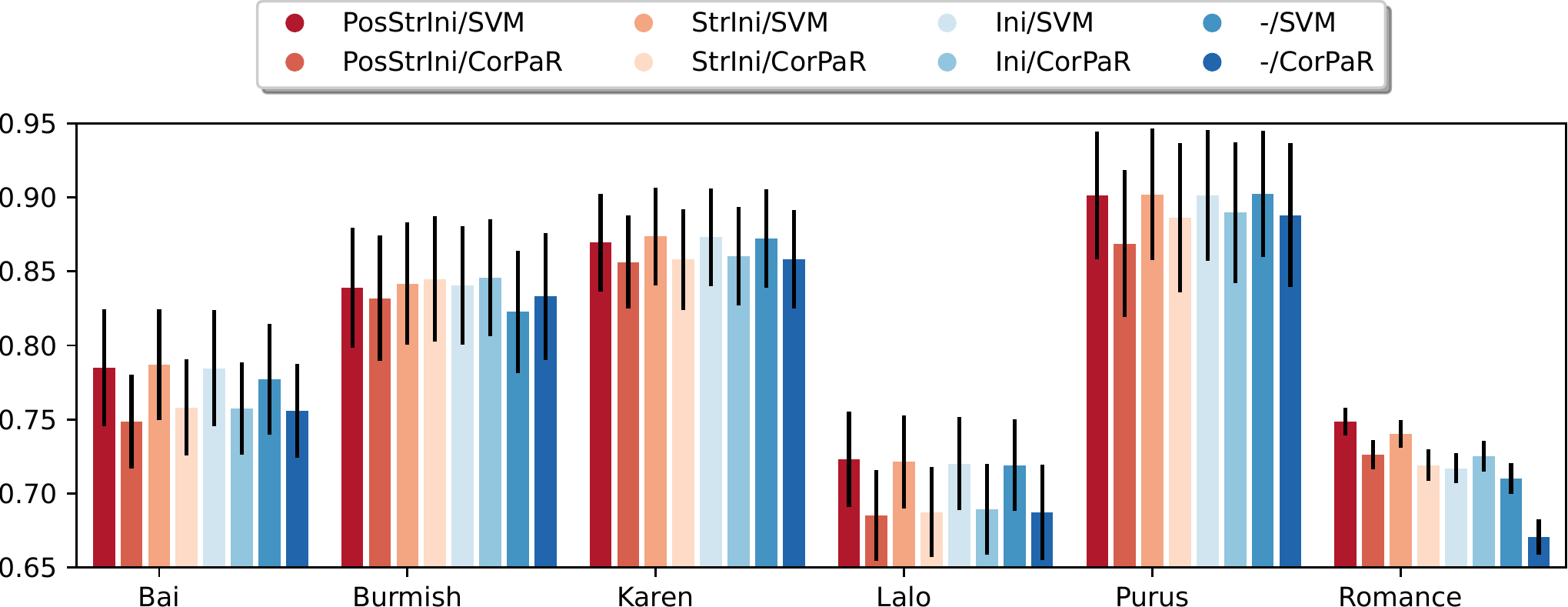}
    \caption{Comparing the results for selected coding techniques and classifiers on individual datasets.}
    \label{fig:fig5}
\end{figure*}

\subsection{Implementation}
The new framework is implemented as part of the LingRex Python package \citep{LingRex} and allows the use of classifiers from the Scikit-Learn Python package \citep{scikit-learn}.

\section{Results}

In order to evaluate the framework, we tested two classifiers, a Support Vector Machine, and the CorPaR classifier (see §~\ref{classifiers}). Furthermore, we tested three different forms of alignment enrichment by coding individual positions (\texttt{Pos}), prosodic structure (\texttt{Str}), as well as whether a sound appears in the beginning or the end (\texttt{Ini}). For each test, we ran 100 trials in which 90\% of the data were used for training and 10\% for evaluation. 
 
\begin{table}[h]
    \centering
    \resizebox{\linewidth}{!}{%
    \begin{tabular}{|l|l|c|c|c|}
\hline
 Classifier   & Analysis   &     ED &    NED &     BC \\\hline
\hline
 SVM          & PosStrIni  & 0.7491 & 0.1598 & 0.8110 \\
 SVM          & PosStr     &\cellcolor{lightgray} 0.7478 &\cellcolor{lightgray} 0.1594 &\cellcolor{lightgray} 0.8115 \\
 SVM          & PosIni     & 0.7701 & 0.1624 & 0.8077 \\
 SVM          & StrIni     & 0.7578 & 0.1601 & 0.8110 \\
 SVM          & Pos        & 0.7685 & 0.1618 & 0.8084 \\
 SVM          & Str        & 0.7681 & 0.1614 & 0.8086 \\
 SVM          & Ini        & 0.7895 & 0.1641 & 0.8061 \\
 SVM          & none       & 0.8059 & 0.1673 & 0.8006 \\\hline\hline
 CorPaR       & PosStrIni  & 0.8503 & 0.1816 & 0.7862 \\
 CorPaR       & PosStr     & 0.8655 & 0.1826 & 0.7854 \\
 CorPaR       & PosIni     & 0.8425 & 0.1802 & 0.7882 \\
 CorPaR       & StrIni     & 0.8402 & 0.1771 & 0.7924 \\
 CorPaR       & Pos        & 0.8836 & 0.1847 & 0.7840 \\
 CorPaR       & Str        & 0.9048 & 0.1851 & 0.7848 \\
 CorPaR       & Ini        &\cellcolor{lightgray} 0.8342 &\cellcolor{lightgray} 0.1763 &\cellcolor{lightgray} 0.7946 \\
 CorPaR       & none       & 0.9379 & 0.1898 & 0.7821 \\
\hline
\end{tabular}}
    \caption{Results for edit distance, normalized edit distance, and B-Cubed F-Scores on all datasets.}
    \label{tab:res1}
\end{table}

Table \ref{tab:res1} shows the results for all eight combinations between the three techniques for
alignment enrichment. As can be seen, the SVM classifier outperforms the CorPaR method, although the
differences are not very large. While the impact of the alignment enrichment techniques on the
results is not very large, we still find that they enhance the results in all SVM trials, while the
raw coding of the position (\texttt{Pos}) leads to lower scores for the CorPaR classifier in our
test set. For the SVM classifier, coding for prosodic structure (\texttt{Str}) and information on
whether a segment occurs at the beginning, in the middle, or the end of a sequence (\texttt{StrIni})
yields the best results with respect to all measures, while \texttt{Ini} coding outperforms the other techniques for the CorPaR classifier. From these results, we can see that alignment enrichment is a promising technique that deserves further exploration, but we do not think that the current codings are the last word on the topic.

Figure \ref{fig:fig5} compares the results for four coding techniques on individual datasets. As can be seem from the figure, the impact of the coding techniques varies quite drastically across datasets. This shows that it would be premature to rule out any of the techniques tested here directly, but rather calls for a careful selection of alignment enrichment techniques dependent on the language family one wants to investigate.

\section{Conclusion}
In this study, we have presented a new framework for supervised phonological reconstruction, which is implemented in the form of a small Python package. The new framework has the advantage of being easy to use, easy to extend, and fast to apply, while at the same time yielding promising results on a newly compiled collection of datasets from three different languages families. Given that our framework can be easily extended, by varying the individual components of the worfklow, 
we hope that it will provide a solid basis for future work on phonological reconstruction, as well
as the prediction of words from cognate reflexes
\citep{Bodt2021,Dekker2021,Beinborn2013,Fourrier2021} in
computational historical linguistics.


\bibliographystyle{acl_natbib}

\begin{thebibliography}{41}
\expandafter\ifx\csname natexlab\endcsname\relax\def\natexlab#1{#1}\fi

\bibitem[{Amigó et~al.(2009)Amigó, Gonzalo, Artiles, and Verdejo}]{Amigo2009}
Enrique Amigó, Julio Gonzalo, Javier Artiles, and Felisa Verdejo. 2009.
\newblock {A} comparison of extrinsic clustering evaluation metrics based on
  formal constraints.
\newblock \emph{Information Retrieval}, 12(4):461--486.

\bibitem[{Anderson et~al.(2018)Anderson, Tresoldi, Chacon, Fehn, Walworth,
  Forkel, and List}]{Anderson2018}
Cormac Anderson, Tiago Tresoldi, Thiago~Costa Chacon, Anne-Maria Fehn, Mary
  Walworth, Robert Forkel, and Johann-Mattis List. 2018.
\newblock \href {https://doi.org/10.2478/yplm-2018-0002} {{A}
  {C}ross-{L}inguistic {D}atabase of {P}honetic {T}ranscription {S}ystems}.
\newblock \emph{Yearbook of the Poznań Linguistic Meeting}, 4(1):21--53.

\bibitem[{Beinborn et~al.(2013)Beinborn, Zesch, and Gurevych}]{Beinborn2013}
Lisa Beinborn, Torsten Zesch, and Iryna Gurevych. 2013.
\newblock Cognate production using character-based machine translation.
\newblock In \emph{{P}roceedings of the {S}ixth {I}nternational {J}oint
  {C}onference on {N}atural {L}anguage {P}rocessing}, pages 883--891.

\bibitem[{Bodt and List(2019)}]{Bodt2019}
Timotheus~A. Bodt and Johann-Mattis List. 2019.
\newblock \href {https://doi.org/10.2218/pihph.4.2019.3037} {{T}esting the
  predictive strength of the comparative method: {A}n ongoing experiment on
  unattested words in {W}estern {K}ho-{B}wa languages}.
\newblock \emph{Papers in Historical Phonology}, 4(1):22--44.

\bibitem[{Bodt and List(2022)}]{Bodt2021}
Timotheus~Adrianus Bodt and Johann-Mattis List. 2022.
\newblock \href {https://doi.org/10.1075/dia.20009.bod} {{Reflex prediction. A
  case study of Western Kho-Bwa}}.
\newblock \emph{Diachronica}, 39(1):1--38.

\bibitem[{Bouchard-Côté et~al.(2013)Bouchard-Côté, Hall, Griffiths, and
  Klein}]{Bouchard-Cote2013}
Alexandre Bouchard-Côté, David Hall, Thomas~L. Griffiths, and Dan Klein.
  2013.
\newblock {A}utomated reconstruction of ancient languages using probabilistic
  models of sound change.
\newblock \emph{Proceedings of the National Academy of Sciences of the United
  States of America}, 110(11):4224–4229.

\bibitem[{Ciobanu and Dinu(2018)}]{Ciobanu2018}
Alina~Maria Ciobanu and Liviu~P. Dinu. 2018.
\newblock {S}imulating language evolution: {A} tool for historical linguistics.
\newblock In \emph{{P}roceedings of the 27th {I}nternational {C}onference on
  {C}omputational {L}inguistics: {S}ystem {D}emonstrations}, pages 68--72.
  Association of Computational Linguistics.

\bibitem[{Cristea et~al.(2021)Cristea, Dinu, Georgescu, Mihai, and
  Uban}]{Cristea2021}
Alina~Maria Cristea, Liviu~P. Dinu, Simona Georgescu, Mihnea-Lucian Mihai, and
  Ana~Sabina Uban. 2021.
\newblock \href {https://aclanthology.org/2021.findings-emnlp.243} {Automatic
  discrimination between inherited and borrowed {L}atin words in {R}omance
  languages}.
\newblock In \emph{Findings of the Association for Computational Linguistics:
  EMNLP 2021}, pages 2845--2855, Punta Cana, Dominican Republic. Association
  for Computational Linguistics.

\bibitem[{de~Carvalho(2021)}]{Carvalho2021}
Fernando~O. de~Carvalho. 2021.
\newblock \href {https://doi.org/10.1086/711607} {A comparative reconstruction
  of proto-purus (arawakan) segmental phonology}.
\newblock \emph{International Journal of American Linguistics}, 87(1):49--108.

\bibitem[{Dekker and Zuidema(2021)}]{Dekker2021}
Peter Dekker and Willem Zuidema. 2021.
\newblock \href {https://doi.org/10.15398/jlm.v8i2.268} {Word prediction in
  computational historical linguistics}.
\newblock \emph{Journal of Language Modelling}, 8(2):295–336.

\bibitem[{Dinu and Ciobanu(2014)}]{Dinu2014}
Liviu Dinu and Alina~Maria Ciobanu. 2014.
\newblock \href
  {http://www.lrec-conf.org/proceedings/lrec2014/pdf/175_Paper.pdf} {Building a
  dataset of multilingual cognates for the {R}omanian lexicon}.
\newblock In \emph{Proceedings of the Ninth International Conference on
  Language Resources and Evaluation ({LREC}'14)}, pages 1038--1043, Reykjavik,
  Iceland. European Language Resources Association (ELRA).

\bibitem[{Forkel et~al.(2021)Forkel, Greenhill, Bibiko, Rzymski, Tresoldi, and
  List}]{PyLexibank}
Robert Forkel, Simon~J Greenhill, Hans-Jörg Bibiko, Christoph Rzymski, Tiago
  Tresoldi, and Johann-Mattis List. 2021.
\newblock \href {https://pypi.org/project/pylexibank} {\emph{PyLexibank. The
  python curation library for lexibank [{Software Library, Version 2.8.2}]}}.
\newblock Zenodo, Geneva.

\bibitem[{Forkel and List(2020)}]{Forkel2020}
Robert Forkel and Johann-Mattis List. 2020.
\newblock \href
  {http://www.lrec-conf.org/proceedings/lrec2020/pdf/2020.lrec-1.864.pdf}
  {Cldfbench. give your cross-linguistic data a lift}.
\newblock In \emph{{P}roceedings of the {T}welfth {I}nternational {C}onference
  on {L}anguage {R}esources and {E}valuation}, pages 6997--7004, Luxembourg.
  European Language Resources Association (ELRA).

\bibitem[{Forkel et~al.(2018)Forkel, List, Greenhill, Rzymski, Bank, Cysouw,
  Hammarström, Haspelmath, Kaiping, and Gray}]{Forkel2018a}
Robert Forkel, Johann-Mattis List, Simon~J. Greenhill, Christoph Rzymski,
  Sebastian Bank, Michael Cysouw, Harald Hammarström, Martin Haspelmath,
  Gereon~A. Kaiping, and Russell~D. Gray. 2018.
\newblock \href {https://doi.org/10.1038/sdata.2018.205} {{C}ross-{L}inguistic
  {D}ata {F}ormats, advancing data sharing and re-use in comparative
  linguistics}.
\newblock \emph{Scientific Data}, 5(180205):1--10.

\bibitem[{Fourrier et~al.(2021)Fourrier, Bawden, and Sagot}]{Fourrier2021}
Cl{\'e}mentine Fourrier, Rachel Bawden, and Beno{\^\i}t Sagot. 2021.
\newblock \href {https://doi.org/10.18653/v1/2021.findings-acl.75} {Can cognate
  prediction be modelled as a low-resource machine translation task?}
\newblock In \emph{Findings of the Association for Computational Linguistics:
  ACL-IJCNLP 2021}, pages 847--861, Online. Association for Computational
  Linguistics.

\bibitem[{Fox(1995)}]{Fox1995}
Anthony Fox. 1995.
\newblock \emph{{L}inguistic reconstruction}.
\newblock Oxford University Press, Oxford.

\bibitem[{Geisler(1992)}]{Geisler1992}
Hans Geisler. 1992.
\newblock \emph{{A}kzent und {L}autwandel in der {R}omania}.
\newblock Narr, Tübingen.

\bibitem[{Gong and Hill(2020)}]{Gong2020}
Xun Gong and Nathan Hill. 2020.
\newblock \href {https://doi.org/10.5281/zenodo.4311182} {\emph{{Materials for
  an Etymological Dictionary of Burmish}}}.
\newblock Zenodo, Geneva.

\bibitem[{Grimm(1822)}]{Grimm1822}
Jacob Grimm. 1822.
\newblock \href {http://arxiv.org/abs/MnsKAAAAIAAJ} {\emph{{D}eutsche
  {G}rammatik}}, 2 edition, volume~1.
\newblock Dieterichsche Buchhandlung, Göttingen.

\bibitem[{Hammarström et~al.(2021)Hammarström, Haspelmath, Forkel, and
  Bank}]{Glottolog}
Harald Hammarström, Martin Haspelmath, Robert Forkel, and Sebastiaon Bank.
  2021.
\newblock \href {http://arxiv.org/abs/https://glottolog.org} {\emph{{G}lottolog
  [{Dataset, V}ersion 4.5]}}.
\newblock Max Planck Institute for Evolutionary Anthropology, Leipzig.

\bibitem[{Jäger(2019)}]{Jaeger2019}
Gerhard Jäger. 2019.
\newblock \href {https://doi.org/10.1515/tl-2019-0011} {{C}omputational
  historical linguistics}.
\newblock \emph{Theoretical Linguistics}, 45(3-4):151--182.

\bibitem[{Jäger et~al.(2017)Jäger, List, and Sofroniev}]{Jaeger2017}
Gerhard Jäger, Johann-Mattis List, and Pavel Sofroniev. 2017.
\newblock \href {http://aclweb.org/anthology/E/E17/E17-1113.pdf} {{U}sing
  support vector machines and state-of-the-art algorithms for phonetic
  alignment to identify cognates in multi-lingual wordlists}.
\newblock In \emph{{P}roceedings of the 15th {C}onference of the {E}uropean
  {C}hapter of the {A}ssociation for {C}omputational {L}inguistics. {L}ong
  {P}apers}, pages 1204--1215, Valencia. Association for Computational
  Linguistics.

\bibitem[{Kondrak(2000)}]{Kondrak2000}
Grzegorz Kondrak. 2000.
\newblock {A} new algorithm for the alignment of phonetic sequences.
\newblock In \emph{{P}roceedings of the 1st {N}orth {A}merican chapter of the
  {A}ssociation for {C}omputational {L}inguistics conference}, pages 288--295.

\bibitem[{Levenshtein(1965)}]{Levenshtein1965}
Vladimir.~I. Levenshtein. 1965.
\newblock {D}voičnye kody s ispravleniem vypadenij, vstavok i zameščenij
  simvolov.
\newblock \emph{Doklady Akademij Nauk SSSR}, 163(4):845--848.

\bibitem[{List(2012)}]{List2012c}
Johann-Mattis List. 2012.
\newblock \href {https://doi.org/10.1007/978-3-642-31467-4_3} {{SCA}:
  {P}honetic alignment based on sound classes}.
\newblock In Marija Slavkovik and Dan Lassiter, editors, \emph{{N}ew directions
  in logic, language, and computation}, pages 32--51. Springer, Berlin and
  Heidelberg.

\bibitem[{List(2014)}]{List2014d}
Johann-Mattis List. 2014.
\newblock \href {https://doi.org/10.1515/9783110720082} {\emph{{S}equence
  comparison in historical linguistics}}.
\newblock Düsseldorf University Press, Düsseldorf.

\bibitem[{List(2017)}]{List2017d}
Johann-Mattis List. 2017.
\newblock \href {http://aclweb.org/anthology/E/E17/E17-3003.pdf} {{A} web-based
  interactive tool for creating, inspecting, editing, and publishing
  etymological datasets}.
\newblock In \emph{{P}roceedings of the 15th {C}onference of the {E}uropean
  {C}hapter of the {A}ssociation for {C}omputational {L}inguistics. {S}ystem
  {D}emonstrations}, pages 9--12, Valencia. Association for Computational
  Linguistics.

\bibitem[{List(2019{\natexlab{a}})}]{List2019a}
Johann-Mattis List. 2019{\natexlab{a}}.
\newblock \href {https://doi.org/10.1162/coli_a_00344} {{A}utomatic inference
  of sound correspondence patterns across multiple languages}.
\newblock \emph{Computational Linguistics}, 45(1):137--161.

\bibitem[{List(2019{\natexlab{b}})}]{List2019e}
Johann-Mattis List. 2019{\natexlab{b}}.
\newblock \href {https://doi.org/10.1515/tl-2019-0016} {{B}eyond {E}dit
  {D}istances: {C}omparing linguistic reconstruction systems}.
\newblock \emph{Theoretical Linguistics}, 45(3-4):1--10.

\bibitem[{List(2021)}]{EDICTOR}
Johann-Mattis List. 2021.
\newblock \href {https://digling.org/edictor} {\emph{{EDICTOR. A web-based tool
  for creating, editing, and publishing etymological datasets [{{Software Tool,
  Version 2.0.0}]}}}}.
\newblock Max Planck Institute for Evolutionary Anthropology, Leipzig.

\bibitem[{List et~al.(2021{\natexlab{a}})List, Anderson, Tresoldi, and
  Forkel}]{CLTS}
Johann-Mattis List, Cormac Anderson, Tiago Tresoldi, and Robert Forkel.
  2021{\natexlab{a}}.
\newblock \href {https://clts.clld.org} {\emph{{C}ross-{L}inguistic
  {T}ranscription {S}ystems [{D}ataset, {V}ersion 2.1.0]}}.
\newblock Max Planck Institute for the Science of Human History, Jena.

\bibitem[{List and Forkel(2022)}]{LingRex}
Johann-Mattis List and Robert Forkel. 2022.
\newblock \href {https://pypi.org/project/lingrex} {\emph{{{L}ing{R}ex:
  {L}inguistic reconstruction with {L}ing{P}y}}}.
\newblock Max Planck Institute for Evolutionary Anthropology, Leipzig.

\bibitem[{List et~al.(2021{\natexlab{b}})List, Forkel, Greenhill, Rzymski,
  Englisch, and Gray}]{List2021PREPRINTd}
Johann-Mattis List, Robert Forkel, Simon~J. Greenhill, Christoph Rzymski,
  Johannes Englisch, and Russell~D. Gray. 2021{\natexlab{b}}.
\newblock \href {https://doi.org/10.21203/rs.3.rs-870835/v1} {Lexibank: {A}
  public repository of standardized wordlists with computed phonological and
  lexical features [{Preprint, Version 1}]}.
\newblock \emph{Research Square}, pages 1--31.

\bibitem[{List et~al.(2016)List, Lopez, and Bapteste}]{List2016g}
Johann-Mattis List, Philippe Lopez, and Eric Bapteste. 2016.
\newblock \href {http://anthology.aclweb.org/P16-2097} {{U}sing sequence
  similarity networks to identify partial cognates in multilingual wordlists}.
\newblock In \emph{{P}roceedings of the {A}ssociation of {C}omputational
  {L}inguistics 2016 ({V}olume 2: {S}hort {P}apers)}, pages 599--605, Berlin.
  Association of Computational Linguistics.

\bibitem[{List et~al.(2021{\natexlab{c}})List, Rzymski, Greenhill, Schweikhard,
  Pianykh, Tjuka, Hundt, and Forkel}]{Concepticon}
Johann-Mattis List, Christoph Rzymski, Simon~J. Greenhill, Nathanael~E.
  Schweikhard, Kristina Pianykh, Annika Tjuka, Carolin Hundt, and Robert
  Forkel. 2021{\natexlab{c}}.
\newblock \href {http://arxiv.org/abs/https://concepticon.clld.org/}
  {\emph{{C}oncepticon. {A} resource for the linking of concept lists
  [{Dataset, V}ersion 2.5.0]}}.
\newblock Max Planck Institute for the Science of Human History, Jena.

\bibitem[{Luangthongkum(2019)}]{Luangthongkum2019}
Theraphan Luangthongkum. 2019.
\newblock {A view on Proto-Karen phonology and lexicon}.
\newblock \emph{Journal of the Southeast Asian Linguistics Society},
  12(1):i--lii.

\bibitem[{Meloni et~al.(2021)Meloni, Ravfogel, and Goldberg}]{Meloni2021}
Carlo Meloni, Shauli Ravfogel, and Yoav Goldberg. 2021.
\newblock \href {https://www.aclweb.org/anthology/2021.naacl-main.353} {Ab
  antiquo: Neural proto-language reconstruction}.
\newblock In \emph{Proceedings of the 2021 Conference of the North American
  Chapter of the Association for Computational Linguistics: Human Language
  Technologies}, pages 4460--4473, Online. Association for Computational
  Linguistics.

\bibitem[{Pedregosa et~al.(2011)Pedregosa, Varoquaux, Gramfort, Michel,
  Thirion, Grisel, Blondel, Prettenhofer, Weiss, Dubourg, Vanderplas, Passos,
  Cournapeau, Brucher, Perrot, and Duchesnay}]{scikit-learn}
F.~Pedregosa, G.~Varoquaux, A.~Gramfort, V.~Michel, B.~Thirion, O.~Grisel,
  M.~Blondel, P.~Prettenhofer, R.~Weiss, V.~Dubourg, J.~Vanderplas, A.~Passos,
  D.~Cournapeau, M.~Brucher, M.~Perrot, and E.~Duchesnay. 2011.
\newblock \href {https://scikit-learn.org/} {Scikit-learn: Machine learning in
  {P}ython}.
\newblock \emph{Journal of Machine Learning Research}, 12:2825--2830.

\bibitem[{Steiner et~al.(2011)Steiner, Stadler, and Cysouw}]{Steiner2011}
Lydia Steiner, Peter~F. Stadler, and Michael Cysouw. 2011.
\newblock {A} pipeline for computational historical linguistics.
\newblock \emph{Language Dynamics and Change}, 1(1):89--127.

\bibitem[{Wang(2004)}]{Wang2004b}
Feng Wang. 2004.
\newblock \emph{{L}anguage contact and language comparison. {T}he case of
  {B}ai}.
\newblock Phd, City University of Hong Kong, Hong Kong.

\bibitem[{Yang(2011)}]{Yang2011}
Cathryn Yang. 2011.
\newblock \emph{{L}alo regional varieties: {Phylogeny}, dialectometry and
  sociolinguistics}.
\newblock {PhD} dissertation, La Trobe University, Bundoora.

\end{thebibliography}

\newpage
\appendix

\onecolumn
\section{Appendix}
\subsection{Source Code and Data}
The new data collection along with the source code and the data needed to replicate the results
reported in this study have been curated on GitHub at
\url{https://github.com/lingpy/supervised-reconstruction-paper} (Version 1.0) and archived with Zenodo (DOI:
\url{https://doi.org/10.5281/zenodo.6426074}). 
\subsection{Table of Results for Individual Datasets}
\subsubsection{SVM}
\begin{tabular}{lrrrrr}
\hline
 DATASET   &   PosStrIni &   StrIni &    Str &    Ini &   none \\
\hline
 Bai       &      0.7848 &   0.7870 & 0.7832 & 0.7846 & 0.7770 \\
 Burmish   &      0.8388 &   0.8418 & 0.8420 & 0.8405 & 0.8226 \\
 Karen     &      0.8696 &   0.8736 & 0.8734 & 0.8731 & 0.8723 \\
 Lalo      &      0.7232 &   0.7214 & 0.7204 & 0.7202 & 0.7191 \\
 Purus     &      0.9011 &   0.9021 & 0.9016 & 0.9013 & 0.9022 \\
 Romance   &      0.7487 &   0.7401 & 0.7310 & 0.7171 & 0.7103 \\
\hline
\end{tabular}

\subsubsection{CorPaR}
\begin{tabular}{lrrrrr}
\hline
 DATASET   &   PosStrIni &   StrIni &    Str &    Ini &   none \\
\hline
 Bai       &      0.7485 &   0.7581 & 0.7560 & 0.7572 & 0.7560 \\
 Burmish   &      0.8319 &   0.8449 & 0.8422 & 0.8458 & 0.8331 \\
 Karen     &      0.8564 &   0.8581 & 0.8614 & 0.8604 & 0.8581 \\
 Lalo      &      0.6852 &   0.6874 & 0.6890 & 0.6893 & 0.6871 \\
 Purus     &      0.8688 &   0.8865 & 0.8730 & 0.8897 & 0.8880 \\
 Romance   &      0.7262 &   0.7192 & 0.6871 & 0.7253 & 0.6705 \\
\hline
\end{tabular}

\label{sec:appendix}

\end{document}